\documentclass[letterpaper]{article} 
\usepackage[submission]{aaai25}  
\usepackage{times}  
\usepackage{helvet}  
\usepackage{courier}  
\usepackage[hyphens]{url}  
\usepackage{graphicx} 
\usepackage{natbib}  
\usepackage{caption} 
\usepackage{algorithm}
\usepackage{algorithmic}
\usepackage{tabularx}
\usepackage{booktabs}
\usepackage{makecell}
\usepackage{multirow}

\usepackage{newfloat}
\usepackage{listings}
\DeclareCaptionStyle{ruled}{labelfont=normalfont,labelsep=colon,strut=off} 
\floatstyle{ruled}
\newfloat{listing}{tb}{lst}{}
\floatname{listing}{Listing}
\def\showauthors@ontrue{}

\author {
    Sangwook Kim\textsuperscript{\rm 1, \rm 2}\thanks{Work done during the internship at LG AI Research},
    Soonyoung Lee\textsuperscript{\rm 2},
    Jongseong Jang\textsuperscript{\rm 2}\thanks{Corresponding author}
}
\affiliations {
    \textsuperscript{\rm 1}Department of Medical Biophysics, University of Toronto\\
    \textsuperscript{\rm 2}LG AI Research\\
    sangwook.kim@mail.utoronto.ca, \{soonyoung.lee, j.jang\}@lgresearch.ai
}
\title{ChatEXAONEPath: An Expert-level Multimodal Large Language Model for Histopathology Using Whole Slide Images}

\usepackage{bibentry}

\begin{document}
\maketitle

\begin{abstract}
Recent studies have made significant progress in developing large language models (LLMs) in the medical domain, which can answer expert-level questions and demonstrate the potential to assist clinicians in real-world clinical scenarios. 
Studies have also witnessed the importance of integrating various modalities with the existing LLMs for a better understanding of complex clinical contexts, which are innately multi-faceted by nature. 
Although studies have demonstrated the ability of multimodal LLMs in histopathology to answer questions from given images, they lack in understanding of thorough clinical context due to the patch-level data with limited information from public datasets. Thus, developing WSI-level MLLMs is significant in terms of the scalability and applicability of MLLMs in histopathology. 
In this study, we introduce an expert-level MLLM for histopathology using WSIs, dubbed as \textbf{\textit{ChatEXAONEPath}}. We present a retrieval-based data generation pipeline using 10,094 pairs of WSIs and histopathology reports from The Cancer Genome Atlas (TCGA). 
We also showcase an AI-based evaluation protocol for a comprehensive understanding of the medical context from given multimodal information and evaluate generated answers compared to the original histopathology reports. 
We demonstrate the ability of diagnosing the given histopathology images using ChatEXAONEPath with the acceptance rate of 62.9\% from 1,134 pairs of WSIs and reports. Our proposed model can understand pan-cancer WSIs and clinical context from various cancer types. We argue that our proposed model has the potential to assist clinicians by comprehensively understanding complex morphology of WSIs for cancer diagnosis through the integration of multiple modalities.
\end{abstract}

%

\section{Introduction}
Understanding and interpreting histopathology images is a significant task in diagnosing cancer. 
However, investigating those images for cancer diagnosis is a difficult and resource-intensive task due to the complex nature of histopathology images. Thus, it is a common sense that the long-term training is required for pathologists to understand the medical contexts of given images. Even after the acquisition of the specialty in histopathology, the guidelines for cancer diagnosis are keep changing due to the discovery of unseen biomarkers or the advent of new cancer types. 

Thankfully, with the development of digital pathology, studies have introduced several computational methodologies for automated cancer diagnosis and validated the efficacy in the real-world clinical scenario. Especially, deep learning has enabled the understanding of complex and unstructured histopathology images, owing to its ability to capture complex patterns in the images. 

However, existing deep learning models for histopathology still remain stand-alone without any interactions of clinicians with the deep learning models, due to the lack of interpretability and the inability of reasoning from the derived results. 
Thus, it is important to develop interaction-based computational methodologies to efficiently utilize those models, leading to better understanding of morphology of histopathology images for the accurate cancer diagnosis. 

\subsection{Advent of LLMs and impact on histopathology}
Large language model (LLM) is a transformer-based deep learning architecture that has revolutionized natural language processing (NLP) by enabling machines to understand, generate, and respond to human languages with a high degree of accuracy.
LLMs significantly impact the medical domain by assisting in the interpretation of complex medical data, especially the ones written in human languages.
LLMs outperform previous NLP models due to their extensive training on large-scale datasets, transformer architecture with self-attention mechanisms, improved contextual understanding, ability to be fine-tuned for specific tasks, and efficient parallelization during training.

Several studies have shown the impact of LLMs in histopathology, proving their ability to accurately interpret complex medical context and generate expert-level answers from the given questions.

Although the development of LLMs in histopathology has shown the potential to become an efficient deep learning assistants for pathologists, LLMs lack in understanding current clinical situations since LLMs can only understand provided textual information, blocking their approach to the non-textual clinical information without human languages.
Thus, considering the multimodal nature in medicine which may not be always interpreted by human generated texts, developing techniques for integrating multiple modalities in medical data is necessary.

\subsection{Advancements of multimodal LLMs}
Since the advent of Contrastive Language-Image Pretraining (CLIP)\cite{radford2021learningCLIP}, studies have demonstrated the effectiveness of intertwining the visionary and textual information owing to the many-to-many relationships between the two different modalities. This correlation also can provide rich inter-modality supervision, leading to improved performance or zero-shot capabilities. In other words, incorporating multiple modalities with various relationships can broaden the representational space, which can improve the understanding of the given data types.
Thus, multimodal LLMs (MLLM) are able to bind different modalities into an integrated representational space, enabling deep learning models to better understand the interaction of various modalities for target tasks in LLMs.

Recent studies have shown remarkable steps towards developing an MLLM-based copilot for pathologists by training the model using large datasets with pairs of histopathology images and corresponding text reports containing cancer types or cancer stages. By connecting the two modalities, the model greatly improved the ability to understand and generate answers based on the given clinical context, which is critical for the real-world applications.

Despite the great improvement in connecting two different modalities for LLMs, it lacks interpreting the whole slide images (WSI) with the provided clinical context due to the limitation of their training datasets mostly consist of pairs of patch-level histopathology images and texts. 
Due to the high-resolution of WSIs, utilizing LLMs with these images is yet to be explored.
Furthermore, unlike natural images that can be collected without specialized knowledge, the development of clinically applicable MLLMs in the medical field, especially building MLLMs for histopathology, requires large-scale clinical datasets that are significantly challenging to obtain. 

Given the difficulty in acquiring extensive datasets for developing MLLMs for medicine, researchers have leveraged publicly available datasets, particularly patch-based pathology images and caption pairs. 
However, these models lack in understanding of thorough clinical context from the given images due to the unpaired data from the public datasets, \textit{i.e.,} a single patch from an unknown original WSI with limited clinical information. This may lead to the potential negative impact of inaccurately capturing clinical information from the WSIs or resulting in hallucination.

\subsection{Contributions}
In this paper, we introduce an MLLM for histopathology that comprehends multimodal inputs and generates responses accurately with the given clinical context from the limited number of datasets. Our proposed method is a \textbf{Chat} system with \textbf{EX}pert-level \textbf{A}I for every\textbf{ONE} and \textbf{Path}ologists, dubbed as \textbf{ChatEXAONEPath}. 
Our proposed model enables better understanding of complex histopathological features and their correlation with textual descriptions, allowing for accurate and contextually relevant generation. 
Our main contributions are threefold.
\begin{itemize}
    \item We present a retrieval-based data generation pipeline, RAIDER, for generating instruction-tuning datasets using 10,094 pairs of WSIs and histopathology reports from The Cancer Genome Atlas (TCGA). 
    \item We fine-tune large language and vision assistant (LLaVA) using the generated pairs of WSIs and reports. For the seamless integration of vision encoder for WSIs, we employ a specialized vision tower for WSIs consisting of patch encoder and aggregator.
    \item We introduce an AI-based evaluation protocol designed to thoroughly comprehend the medical context from provided multimodal information, evaluating the generated answers with seven criteria. We demonstrate that our model can accurately answer the given questions, \textbf{achieving the acceptance rate of 62.9\% using the test dataset}.
\end{itemize}

\section{Related Works}
\subsection{Vision Language Foundation Models in Medicine}
Advancements in the field of multimodal deep learning have enabled the development of vision-language models in medicine. These models are able to link various medical modalities with corresponding text information \cite{acosta2022multimodal}. The advent of vision-language models \cite{liu2024visual-llava, yu2022coca} interconnecting the two modalities for language processing boosted the understanding relationship of complex medical images and medical terminologies \cite{li2024llava}.

This vision-language relationship provides additional supervision across different modalities, which can also generate contextually relevant descriptions by formulating a joint representation space across different modalities \cite{gao2024medbind}. This allows zero-shot predictions which have shown potential in fields like radiology, dermatology, and pathology, by improving diagnostic accuracy and aiding clinicians in interpreting complex medical data through NLP \cite{tu2024towards-medpalmM, yang2024advancing}.

\begin{table*}[ht!]
\centering
\begin{tabular}{l|llllll}
\toprule
\textbf{Model} & \textbf{Resolution} & \textbf{Function} & \textbf{Evaluation} & \textbf{$\#$ Datasets} & \textbf{Pancancer} & \textbf{Date} \\
\midrule
\makecell[l]{Med-PaLM M \\ \cite{tu2024towards-medpalmM}} & Patch & VQA & \makecell[l]{Generation \\ (BLEU-1, F1)} & 32,799 & True & 2023-07 \\
\midrule
\makecell[l]{CONCH \\ \cite{lu2024visualcaption1}} & Patch & Captioning & \makecell[l]{Zero-shot, \\ Downstream} & 1.17M & True & 2024-02 \\
\midrule
\makecell[l]{PathAsst \\ \cite{sun2024pathasst}} & Patch & Captioning & \makecell[l]{Zero-shot, \\ Generation} & \makecell[l]{207K (\textit{VL.}) \\ + 180K (\textit{Inst.})} & True & 2024-02 \\
\midrule
\makecell[l]{PathChat \\ \cite{lu2024multimodalpathchat}} & Patch & \makecell[l]{Conversation} & \makecell[l]{Human \\ -evaluation} & \makecell[l]{1.18M (\textit{VL.}) \\ + 457K (\textit{Inst.})} & True & 2024-05 \\
\midrule
\makecell[l]{PA-LLaVA \\ \cite{dai2024pallava}} & Patch & \makecell[l]{Conversation} & \makecell[l]{Zero-shot, \\ Generation} & \makecell[l]{0.8M (\textit{VL.}) \\ + 35,543 (\textit{Inst.})} & True & 2024-08 \\
\midrule
\makecell[l]{PRISM \\ \cite{shaikovski2024prism}} & WSI & \makecell[l]{Captioning} & \makecell[l]{Generation, \\ Downstream} & 587K WSIs & True & 2024-05 \\
\midrule
\makecell[l]{PathAlign \\ \cite{ahmed2024pathalign}} & WSI & \makecell[l]{Captioning} & \makecell[l]{Generation, \\ Downstream} & \makecell[l]{350K WSIs}  & \makecell[l]{True} & 2024-06 \\
\midrule
\makecell[l]{WSI-VQA \\ \cite{chen2024wsivqa}} & WSI & \makecell[l]{VQA} & \makecell[l]{Generation, \\ Downstream} & \makecell[l]{977 WSIs + \\ 8672 QA}  & \makecell[l]{False \\ (BRCA)} & 2024-07 \\
\midrule
\midrule
\makecell[l]{\textbf{ChatEXAONEPath-v1} \\ (Ours)} & WSI & Conversation & AI-based evaluation & 10,094 & True & 2024-08 \\
\midrule
\makecell[l]{\textbf{ChatEXAONEPath-v2/v3} \\ (Ours)} & WSI & Conversation & AI-based evaluation & 69,544 & True & 2024-08 \\
\bottomrule
\end{tabular}
\caption{The overview of vision-language models in histopathology. VQA stands for visual question-answering task. WSI stands for whole-slide image. \textit{VL.} stands for vision-language alignment, while \textit{Inst.} stands for the instruction datasets. Downstream tasks for each study may vary from subtype classification to survival analysis. The metrics of generation tasks are mostly NLG (natural language generation) metrics (\textit{e.g.}, BLEU or ROUGE).}
\label{table:model_comparison}
\end{table*}
\subsection{MLLMs in Histopathology}
Histopathology, which involves the microscopic examination of tissue samples, presents challenges due to the intricate and multifaceted nature of the data. 
Multimodal approaches, which combine visual information from WSIs with textual data from pathology reports, have emerged as promising solutions to address these challenges. 
By leveraging the strengths of both image analysis and NLP, these multimodal models aim to provide comprehensive understanding of histopathological data. 
Recent research trend has focused on integrating WSIs and textual information to develop robust models that can interpret and analyze complex contexts effectively.

Although existing multimodal LLMs in histopathology have proven the effectiveness of intertwining the two different medical modalities, the generality of these models is still limited due to the scale of high-quality training datasets. 
Thus, studies have released public pathology datasets with pairs of histopathology images and text reports extracted from twitter or medical textbooks \cite{huang2023visualtwitter, gamper2021multipletextbooks}.
A study made a progress towards building question and answer sets for pathology images, which can be utilized to fine-tune LLMs for generative purposes \cite{ikezogwo2024quilt}. This extends earlier studies by enabling the MLLMs available for the dialogue with human experts.

With the advancement of zero-shot capability of MLLMs, studies have presented a potential of MLLMs in generative tasks such as captioning or visual question answering (VQA) from the given histopathology images and text as inputs \cite{lu2024visualcaption1, lu2023visualcaption2, naseem2023kcaption3, naseem2022visioncaption4, he2020pathvqacaption5}. Especially, a remarkable study introduced a multimodal generative copilot for computational pathology, PathChat \cite{lu2024multimodalpathchat}. PathChat is a multimodal chat system which can assist human pathologists by answering diverse types of questions regarding the given image. PathChat was trained on 1.18 million image-caption pairs and 456,916 question and answer sets. This large-scale dataset allows PathChat to capture robust patterns of histopathology images and textual information. 
PathChat showed a significant achievement in generating various multimodal questions and answer sets.

Although these datasets are useful in terms of relating histopathology images with corresponding text, they are limited in capturing the overview understanding of the entire images. In the meantime, these models are mostly comprised of patch-level pathology images that are sub-regions extracted from WSIs where primary diagnosis can be observed limiting the interpretation of the overview of the entire clinical context.

A study presented a slide-level MLLM for histopathology, called PRISM \cite{shaikovski2024prism}, which was trained on around 587,000 WSIs and reports pairs. The proposed model can perform a WSI-level generative captioning task and allow for zero-shot prediction of various histopathological tasks. However, this study is still limited in generating interactive answers which need further development.

In this study, we introduce a MLLM for histopathology, ChatEXAONEPath, which can perform multimodal generative tasks from the limited number of datasets with WSIs-text pairs. We extensively validated our model using open-source LLM to fairly and efficiently evaluate the generative ability of MLLM in pathology.
\section{Methodology}
\begin{figure*}[ht!]
\centering
\includegraphics[width=0.9\textwidth]{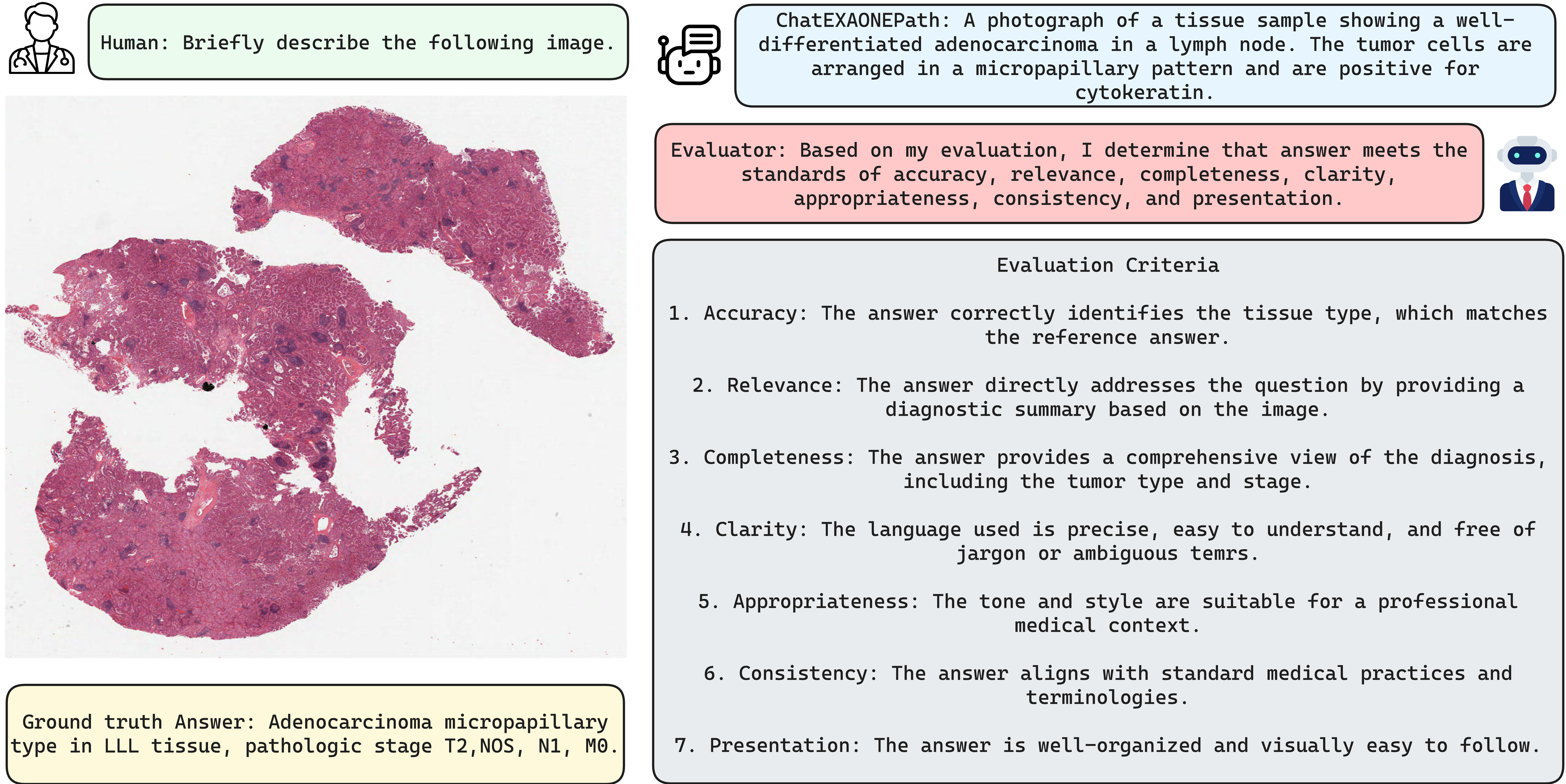}
\caption{A qualitative result and evaluation from the test dataset. ChatEXAONEPath answered from the given whole slide image and question. The evaluator AI model generated reasons to accept the best answer, and made a final decision.}
\label{fig:result}
\end{figure*}
\subsection{Model Development}
In this section, we explain the process of developing ChatEXAONEPath. The development consists of two phases, (1) vision-language alignment and (2) instruction fine-tuning.
The first phase of the model training, vision-language alignment, integrates the representations of two different modalities into one representational space by training vision stream. 
The second phase is the instruction tuning stage where we fine-tune LLM with the aligned modality encoders from the first phase. We use generated instruction datasets for fine-tuning language stream in this second phase, which will be explained in the following subsections.

\subsubsection{Backbone}
We present three primary backbone components of ChatEXAONEPath. We design the overall structure similar to PathChat \cite{lu2024multimodalpathchat}. Here, we employ an additional module for the seamless integration of WSIs.

\textbf{Vision Tower} We introduce a vision tower to process visionary information of WSIs in our proposed MLLM. The vision tower consists of two modules for the patch-based WSI processing, (1) patch encoder and (2) patch aggregator. 

\textbf{Patch Encoder} To process WSIs, we extract patches from the input WSI. The size of the extracted patches is 256 x 256 with the 0.5 microns per pixel from the 20x zoomed WSIs.
We utilize a state-of-the-art patch encoder, EXAONEPath \cite{yun2024enhancingEXAPath}, a robust patch-level foundational model for histopathology images. The structure of EXAONEPath consists of a ViT \cite{dosovitskiy2020image}, which was trained on roughly 2.8 million patches from the TCGA and Genotype-Tissue Expression (GTEx) datasets.
We employ pretrained model weights from EXAONEPath which was trained via self-supervised contrastive learning manner. We freeze the weights of EXAONEPath in the vision tower, treating the patch encoder as a patch-level feature extractor. The detailed information of the patch encoder can be found in the original paper. 

\textbf{Patch Aggregator} To integrate patch-level embeddings with a single WSI-level vision embedding, we adopt clustering-constrained-attention multiple-instance learning (CLAM) \cite{lu2021dataclam}. \textbf{C}LAM-\textbf{B}ased \textbf{P}atch \textbf{A}ggregator (CBPA) employs attention-based learning to pinpoint sub-regions with high significance, extracting a single WSI representative embedding.
CBPA consists of a fully-connected layer followed by \textbf{G}ated \textbf{AT}tention \textbf{N}etwork (GATN). GATN is a module for aggregating patch-level embeddings, which employs two different attention modules consisting of a linear layer followed by an activation function, one with Tanh and the other with Sigmoid function. With the two different activation functions, GATN can capture robust aggregated features from patch embeddings. We multiplied the two outputs from each attention module, and loaded the multiplied output to a linear layer, getting the final output of GATN.

In this study, we separately developed CBPA to embed genetic information into our proposed MLLM. We pretrained CBPA using contrastive learning between WSI feature from CBPA and corresponding RNA sequencing information from selected genes using high-dimensional weighted gene correlation network analysis \cite{langfelder2008hdwgcna, kim2024pancancergene}. With the integrated genetic information, we efficiently built multimodal aggregator encompassing patch-level histopathology images with corresponding RNA sequencing information to aggregate patches in a robust way.

\textbf{Vision projector} Similar to PathChat, we devise a vision projector which projects vision embeddings from the vision tower onto the textual embeddings.
Vision projector consists of attention pooler \cite{yu2022coca} followed by layer normalization and a linear projector (Linear layer – GeLU – Linear layer).
Attention pooling technique pools a single image embedding from multiple image tokens. Although the input of the attention pooler module has a single image embedding feature, we kept the original structure of the attention pooler. We observed that the attention pooler in the vision projector can smoothly project the image embeddings by weighting the vision embedding using self-attention.

\textbf{LLaMA2}
Large Language Model Meta AI (LLaMA) \cite{touvron2023llama2} is an open-source LLM developed by Meta. We utilize pretrained LLaMA to process textual data. We used LLaMA2:7B-Chat model from the collection of LLaMA2 models, which was fine-tuned for dialogue-based applications.
In this study, especially, we employ LLaVA-style MLLM which is an vision-language MLLM with LLaMA as a language backbone. Unlike LLaMA, LLaVA gets both text and image inputs by concatenating the embeddings from different modalities. With the concatenated embeddings as inputs, LLaVA can understand the linear relationship between image and text embeddings.

\subsection{Phase 1: Vision-language alignment}
\begin{figure}[t]
\centering
\includegraphics[width=0.95\columnwidth]{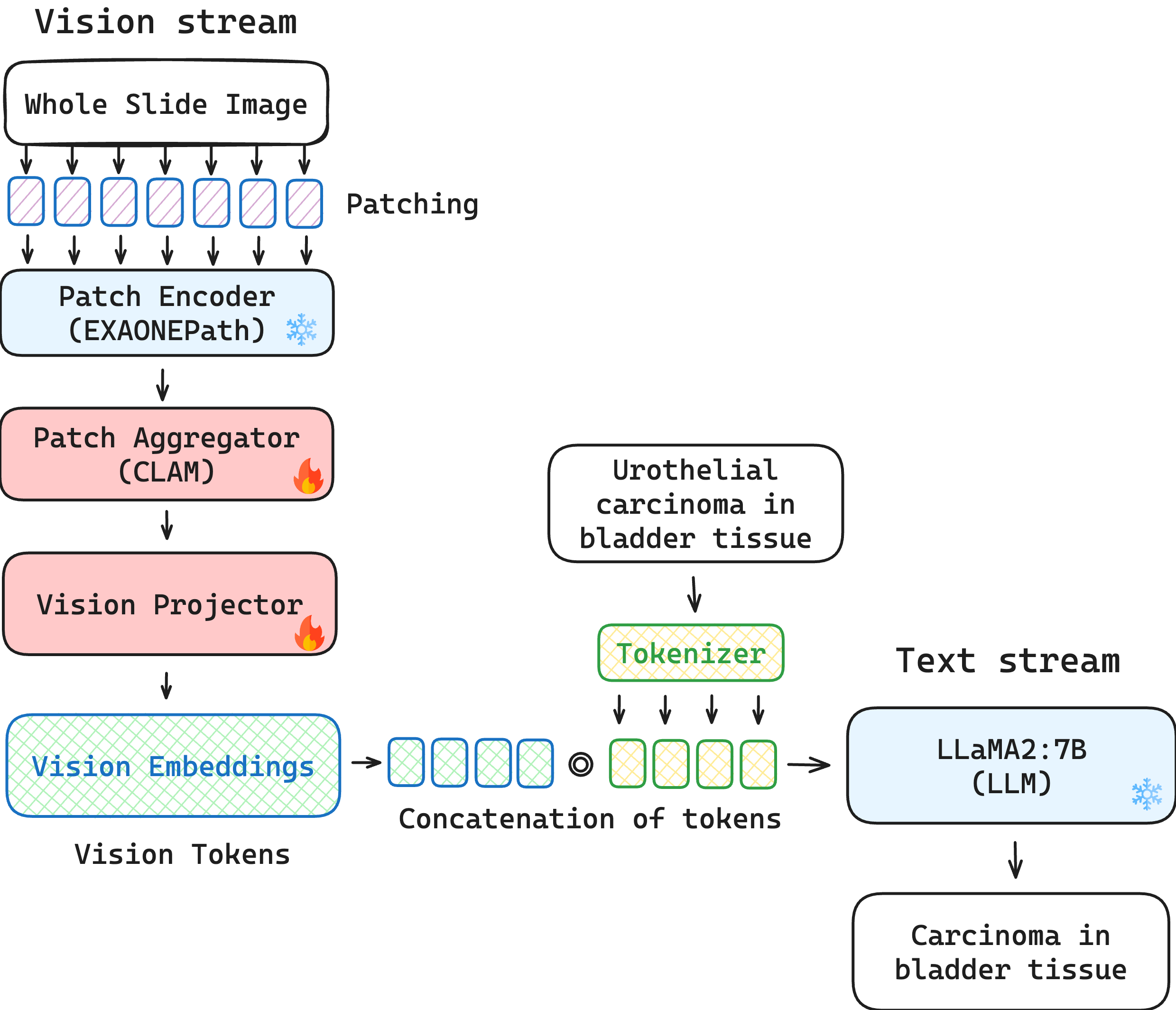}
\caption{The overview of the structure of ChatEXAONEPath for training phase 1, vision-language alignment of whole slide images and corresponding captions.}
\label{fig:phase1}
\end{figure}
Vision-language alignment requires multimodal pairs with histopathology images with the corresponding captions. We use pairs of WSIs and corresponding reports for this phase.
The purpose of the alignment is to pretrain the vision tower and vision projector to generate projected visual embeddings, which are aligned with textual embeddings.
Vision projector gets the output of the vision tower from the dimension of (batch size $\times$ 512), and projects the input vision embedding into the dimension of (batch size $\times$ 4096). The output dimension of our vision projector is 4096 to match the dimension of text and vision input embeddings.

Lastly, the final output of the vision projector is concatenated with text embeddings, which are the inputs to the language model, LLaMA. LLaMA generates captions in an auto-regressive manner generating captions from the given input multimodal embeddings.
We train the pretraining phase using cross-entropy loss between the logits of each token and the corresponding label token. This is similar to training VQA models that are generating series of tokens by predicting the token-wise probability.
It is important to note that only weights of the vision tower are trainable during the first pretraining phase, while the language stream is frozen, focusing on the vision and text alignment. The graphical overview of phase 1 is shown in Figure \ref{fig:phase1}.

\subsection{Phase 2: Instruction Tuning}
\begin{figure}[t]
\centering
\includegraphics[width=0.95\columnwidth]{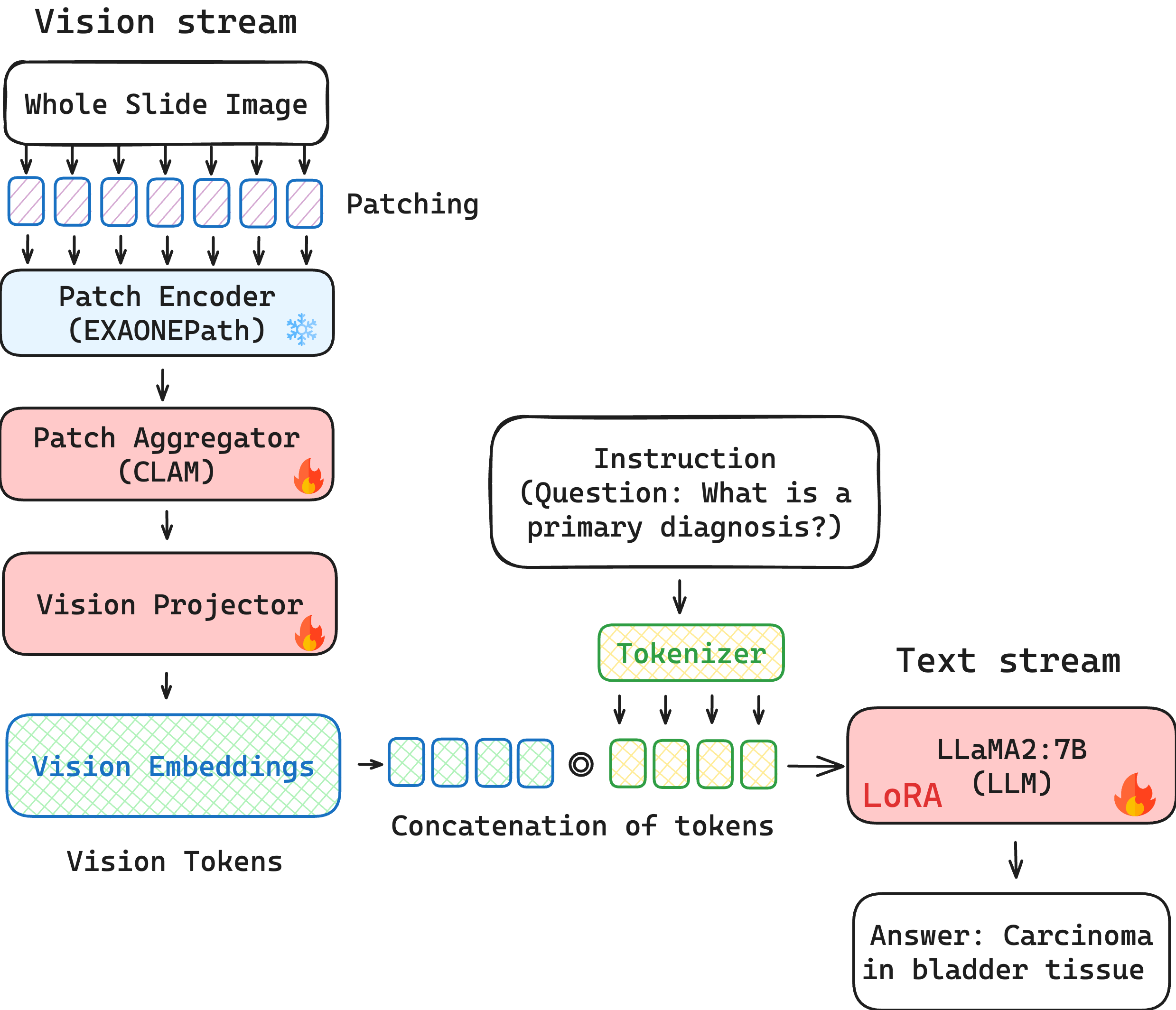} 
\caption{The overview of the structure of ChatEXAONEPath for training phase 2, instruction tuning of LLaMA2:7B model for generating answers from the given input question and the input whole slide image. Compared to the phase 1, the weights of linear projection layers are fine-tuned using low-rank adaptation technique (LoRA).}
\label{fig:phase2}
\end{figure}

In phase 2, we train LLaMA to generate answers based on the questions and instructions by unfreezing the weights of LLaMA. In the instruction tuning phase, we use instruction datasets consist of question and answers with proper instructions. Herein, instead of fine-tuning the entire weights of the language model, we utilized Low-Rank Adaptation of Large Language Models (LoRA) \cite{hu2021lora}. With LoRA, we fine-tuned LLaMA in a computationally efficient way while preserving the performance of LLaMA. We visualize the overview of phase 2 in Figure \ref{fig:phase2}.

Unlike the pretraining phase, we employed a system prompt to provide detailed context and instructions to the language model. By providing an appropriate instruction, we observed a significant improvement in understanding and generation performance. We follow the basic format of prompts and dialogues of LLaMA2.
Similar to phase 1, we use cross-entropy loss for phase 2, training the model to accurately generate responses based on the instructions.

\subsection{Retrieval-augmented Dataset Generation}
Retrieval Augmented Generation (RAG) \cite{lewis2020retrievalRAG} is a technique that combines retrieved knowledge from external sources with a generative model to produce more accurate and contextually relevant responses. 
Considering the heterogeneity of information in the reports, RAG has advantages over the simple contextualization by providing the LLM with concise information and alleviating the hallucination.
We utilize RAG to generate instruction datasets from the original text. We call this process \textbf{R}etrieval-\textbf{A}ugmented \textbf{I}nstruction \textbf{D}ataset g\textbf{E}ne\textbf{R}ation (\textbf{RAIDER}). We employed RAIDER to generate accurate dialogue datasets using text reports.
RAIDER consists of three steps: (1) optical character recognition (OCR), (2) building a vector database, and (3) generation of answers.

We first convert the original reports with portable document format into plain text using publicly available OCR called Doctr \cite{doctr2021ocr}.
We build a vector database containing textual information from the extracted documents. Here, we use Chroma DB to build a vector database where we can store chunks of text for the retrieval.
We then find the most relevant chunks from the vector database by calculating the semantic similarity, cosine distance, between the given question and the text of each chunk.

For the final step, we formulate the input text for an open-source LLM to generate answers based on the given extracted chunks as context. Here, we write a system prompt with a detailed context which can assist the LLM to generate accurate answers with the given retrieved text. We utilize state-of-the-art publicly-available LLM to generate answers, LLaMA3.1:70b-instruct \cite{dubey2024llama3}, which was instruction-tuned from the original LLaMA3.1.
Note that the prompts used in our study is deeply inspired by the one used in PRISM \cite{shaikovski2024prism}.

\textit{Prompt: You are a pathology lab assistant. You are given an unstructured pathology report describing a tissue sample of whole slide image. Follow these instructions carefully: 1. Extract a detailed summary of the diagnosis and the examined tissue from the report in a sentence under 10 words. 2. Do not mention any cm or mm measurements. 3. Do not mention any arabic or roman numerals. 4. Please give a complete and concise answer for the question. Answer the question based only on the provided context: \{context\} Question: \{question\}.}
\section{Experiments}
\subsection{Dataset}
\begin{table}[t]
\centering
\begin{tabular}{l|ll}
\toprule
\textbf{Version} & \textbf{Method} & \textbf{Num Dataset} \\
\midrule
Dataset-v1 & GPT-4o & 10,094 \\
\midrule
Dataset-v2 & \makecell[l]{LLaMA3.1:70b-instruct \\ + RAIDER} & 69,544 \\
\bottomrule
\end{tabular}
\caption{Overview of two distinct versions of datasets. RAIDER stands for retrieval-augmented instruction dataset generation.}
\label{table1:dataset}
\end{table}
Our study employed The Cancer Genome Atlas (TCGA) pan-cancer histopathology dataset. This publicly available dataset provides diverse digitized WSIs from multiple cancer types, accompanied by comprehensive clinical data. 
The dataset enables correlation of histological features with genomic alterations across various cancers, making it an ideal dataset for computational pathology research.
We used a total of 10,094 WSIs and corresponding reports from the dataset. 
The dataset includes high-resolution WSIs, which are histopathology images representing diverse cancer types such as squamous cell carcinoma, and lung adenocarcinoma. 
In our study, we built two different datasets for WSIs and report pairs extracted from the TCGA dataset. To investigate the impact of these datasets in building MLLM, we trained two different models based on the two distinct versions of the datasets.
We present the overview of the datasets in Table~\ref{table1:dataset}. We used a split of 8,960 train and 1,134 test pairs. For the dataset-v1, the total number of WSIs and report pairs are the same as the original datasets, while the number of pairs in the dataset-v2 is 69,544. 
\subsubsection{Dataset-v1 (GPT-4o)}
Dataset-v1 is a dataset with LLM-generated responses using GPT-4o, requesting the model to generate responses based on the given TCGA reports and prompts. GPT-4o is a variant of the GPT-4 model understanding multimodal pairs of data inputs with human language.GPT-4o can generate coherent and contextually relevant responses across a wide range of topics, including medical histopathology.
The prompt to generate responses from GPT-4o is similar to the prompt used in PRISM \cite{shaikovski2024prism}.
We provided the entire contents from each report to GPT-4o as context, and allowed GPT-4o to generate responses based on the context.
For dataset-v1, we generated 10,094 reprsentative captions from the human-generated reports for WSIs, where the primary diagnosis and noticeable histopathologic features are recorded.
\subsubsection{Dataset-v2 (LLaMA3.1:70b)}
We built dataset-v2 using RAIDER, which is based on LLaMA3.1:70b-instruct.
The motivation behind developing dataset-v2 is to show the feasibility of general purpose open-source models to generate large-scale instruction tuning datasets. 
We intended to improve the MLLM performance by augmenting the original reports.
Additionally, using RAIDER allows for more accurate and robust generation of the datasets, as the original reports may contain information non-relevant to the given prompts. 
Furthermore, RAIDER can mitigate the risk of hallucination—a common side effect of using LLMs—by selectively retrieving relevant information from the entire context and excluding irrelevant content. As a result, we generated 69,544 pairs of instruction datasets.

\subsection{Training Details}
For model development, we used a batch size of 128 for phase 1 and 64 for phase 2. 
We used four NVIDIA-A100-40GB GPUs and distributed the mini-batch equally across the four GPUs. During phase 2, we utilized gradient accumulation with a step size of 2, resulting in an effective batch size of 128 (64 $\times$ 2).
Considering the number of training datasets is different for both versions, we trained the model for 2 epochs and 4 epochs using the dataset-v1 and v2, respectively. We selected the final model after the end of training without any validation steps.
For all datasets and phases, we used the cosine learning rate scheduler with the warm up ratio of 0.03, while keeping the peak learning rate to 2 $\times$ $10^{-3}$ for phase 1, and 2 $\times$ $10^{-5}$ for phase 2. We employed Adam optimizer \cite{kingma2014adam}. 
We used PyTorch ver. 2.0.1 \cite{paszke2019pytorch} for the entire implementation. Our codes are deeply inspired by open-source codes from the PathChat \cite{lu2024multimodalpathchat}.

We adopted LoRA for training LLaMA in phase 2, making all of the projection linear layers in LLaMA (e.g., query, key, output projection layers, etc.) trainable during model training while keeping the rest of the model weights frozen. We set the rank of the LoRA to 64, alpha to 16, and the dropout to 0.05. We used PEFT library \cite{peft} to convert the projection layers in LLaMA to LoRA-enabled trainable modules.
We set the hyperparameters for the generation with the temperature of 0.7, top-k of 50 for top-k filtering, the top-p probability of 0.95, and 128 for the number of max new tokens for generation.
\begin{figure*}[ht!]
\centering
\includegraphics[width=0.95\textwidth]{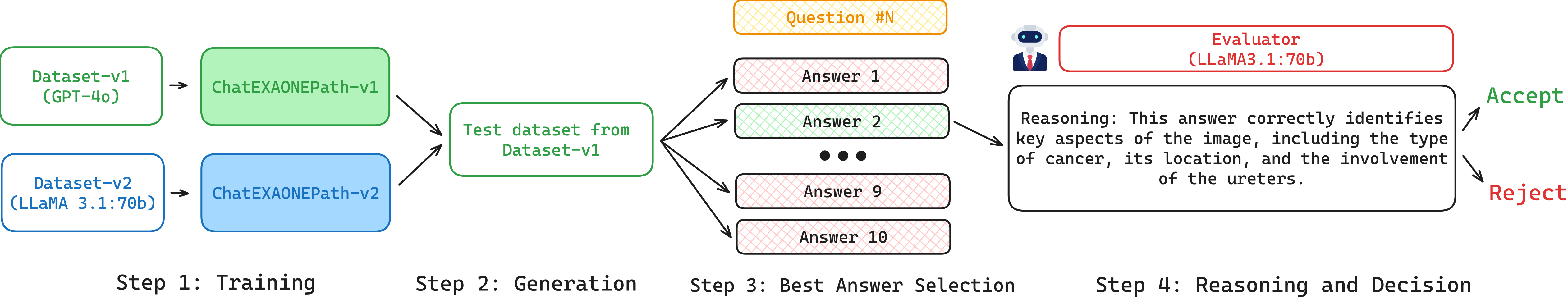} 
\caption{Overview of AI-based evaluation process. After then end of the training, we generated 10 answers per each question, which are then evaluated by the evaluator LLM to make a decision of acceptance or rejection.}
\label{fig:ai_evaluation}
\end{figure*}
\subsection{Evaluation}
We validated the performance of our proposed model by adopting an AI-based evaluation using a well-established LLM, instruction-tuned LLaMA3.1:70b-instruct. We call this well-established LLM an evaluator.
To accurately evaluate the generated answers from the given question, we generated 10 answers for each question and asked the evaluator to select the best answer among the answers.
After choosing the best answer, we let the evaluator decide whether to accept or reject the chosen best answer, innately making the whole evaluation process a multi-task performer. We provide a detailed prompt to the evaluator model during the evaluation process. 
We specifically asked the evaluator to perform the reasoning steps during the decision process by using chain-of-thought (CoT) prompting. This way, we can explicitly interpret the evaluation process behind the evaluator, making the process more valid.
We calculate the acceptance rate (number of accepted responses $/$ a total number of questions in the test dataset), which is a primary evaluation metric for the model performance. 

\begin{table}[t]
\centering
\begin{tabular}{l|ll}
\toprule
\textbf{Version} & \textbf{\makecell[l]{\# Accept}} & \textbf{\makecell[l]{Acceptance \\ Rate}} \\
\midrule
ChatEXAONEPath-v1 & 617 & 54.41\% \\
\midrule
ChatEXAONEPath-v2 & 486 & 42.86\% \\
\midrule
ChatEXAONEPath-v3 & 713 & \textbf{62.87\%} \\
\bottomrule
\end{tabular}
\caption{Acceptance rate of generated answers from the given prompts and questions by the two models. The total number of test dataset is 1,134.}
\label{table2:results}
\end{table}
\subsection{Experiment Results}
We conducted an experiment to generate answers using the two developed models with two distinct datasets. We call these models ChatEXAONEPath-v1 (CEXP-v1) and ChatEXAONEPath-v2 (CEXP-v2), which are ChatEXAONEPath trained with dataset-v1 and dataset-v2. We also trained ChatEXAONEPath-v3 (CEXP-v3) which uses pre-trained EXAONEPath without Macenko normalized images, but still using the dataset-v2.
In this subsection, we showcase the quantitative results and qualitative outputs of generated responses from the two developed models by comparing the responses from each model.
Lastly, we qualitatively analyze the validity of the evaluation generated by the evaluator AI model.

\subsubsection{Quantitative evaluation}
We calculated the acceptance rate of the generated responses based on the evaluations from the evaluator LLM model. We compared the acceptance rate of the two models.
The acceptance rate was calculated based on the test datasets from dataset-v1, which ask for the primary diagnosis or representative diagnostic evidence from the given images.
As shown in Table \ref{table2:results}, the acceptance rate of CEXP-v1 is 54.41\% (617$/$1,134), while the acceptance rate of CEXP-v2 is 42.86\% (486$/$1,134). 
From the experimental results, we observed that the acceptance rate of the CEXP-v1 is higher than that of CEXP-v2. 

These unexpected results show that the MLLM performance may not significantly be related to the quantity of datasets which were augmented without scaling the number of the original pairs. 
LLM-based augmentation of the original vision-text pairs may cause imbalanced alignment of WSIs and textual information.
Thus, the representational space of vision embeddings may not formulate valid correlation with the clinical information written in the limited number of reports.
In the meantime, we foresee that utilizing meaningful pairs of WSIs and text reports may result in better performance for developing MLLMs.

\subsubsection{Evaluation of generated answers and interpretations}
We qualitatively evaluated the generated responses by comparing the two responses from two distinct models.
We also qualitatively validated the evaluator's reasoning based on the evaluation criteria behind the decision.
This analysis is significant in terms of the reliability of the evaluator model, since we generated 10 responses and left the evaluator to choose the best answer out of the generated answers. 
In Figure \ref{fig:result}, we showcase a result of the generated report and reasons from the evaluator. The overview of the evaluation process is shown in Figure \ref{fig:ai_evaluation}. 

We generated an interpretable reasoning procedure of the evaluator model by providing a system prompt with seven different evaluation factors: accuracy, relevance, completeness, clarity, appropriateness, consistency, and presentation. 
The system prompt for the evaluator will be presented in the supplementary material. 

We experimentally showed that the AI evaluator considers all seven standards independently and equally, so that the interpretation can be comprehensive and solid.
In particular, despite using the AI evaluator without medical specialty, the interpretation correctly captures the type or location of tumor which are significant factors when evaluating the generated answers.

However, we observed that reasons for accepting or rejecting an answer are sometimes inaccurate. For example, the model tries to be overly strict about trivial constraints such as length restriction (\textit{e.g.}, generating less than 15 words), although the generated responses include valid diagnosis. 
Furthermore, once the LLM evaluator decides to reject the response at the very beginning, the evaluator provides incorrect or manipulated reasons and treats all seven criteria as a single criterion resulting in generating non-reasonable interpretations.

Although we provided a detailed prompt to the evaluator, the responses and interpretations are somewhat unstable, resulting in incorrect decision making.
This implies that in the worst case, the acceptance rate may be non-trustworthy - leaving the whole AI-based evaluation system incorrect.
Thus, we argue that the false interpretations from the AI evaluator may strictly limit the practicality of the developed MLLMs. 
We leave this issue as our future work.
\section{Conclusion}
In this study, we developed a WSI-level MLLM for human histopathology. 
To acquire large-scale multimodal datasets, we designed RAIDER to scale the raw TCGA WSIs and report pairs, resulting in robust vision-language pairs for training MLLMs. 
We also introduced an interpretable AI-based evaluation system to select the best-generated responses based on the reasoning steps of the evaluator AI model. 
We evaluated the performance of the developed models, trained with two distinct versions of datasets, by calculating the acceptance rate of the generated answers based on the given instruction. 

Despite the contributions, our study is not without limitations. 
First, our experiments showed that training MLLMs with RAIDER-augmented datasets does not always result in improved generation performance. 
We suggest that scaling only a single modality from multimodal dataset pairs may lead to less accurate generative performance. 
Moreover, our AI-based qualitative evaluation system heavily relies on the quality of the ground truth textual information, which remains limited in its unimodal textual understanding of the given answers. 
Therefore, we envision future work involving a multimodal evaluator to validate the generated responses, which can provide multimodal reasoning steps that are efficiently interpretable by human pathologists.

\bibliography{aaai25}
\bigskip
\pagebreak
\setcounter{table}{0}
\setcounter{figure}{0}
\renewcommand{\thetable}{A\arabic{table}}
\renewcommand{\thefigure}{A\arabic{figure}}
\onecolumn
\section{ChatEXAONEPath: A Dataset Perspective of Developing an Expert-level
Multimodal Large Language Model for Histopathology Using Whole Slide Images}
\bigskip
\section{Technical Appendix}
\bigskip

\begin{figure*}[ht!]
\centering
\includegraphics[width=0.9\textwidth]{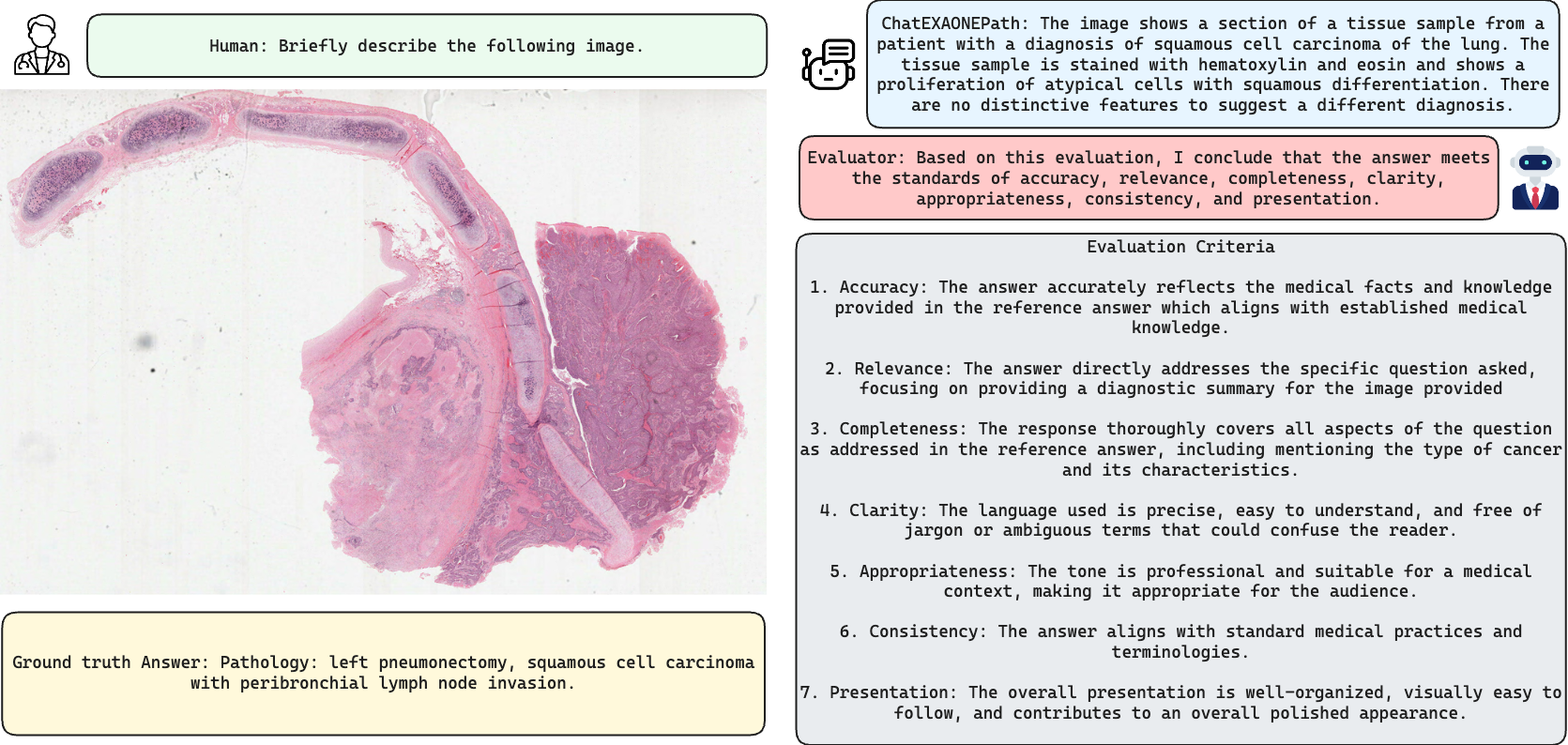}
\caption{A qualitative result and evaluation from the test dataset. ChatEXAONEPath-v3 answered from the given whole slide image and question. The evaluator AI model generated reasons to \textbf{accept} the best answer, and made a final decision.}
\label{fig:result_sup_1}
\end{figure*}

\begin{figure*}[ht!]
\centering
\includegraphics[width=0.9\textwidth]{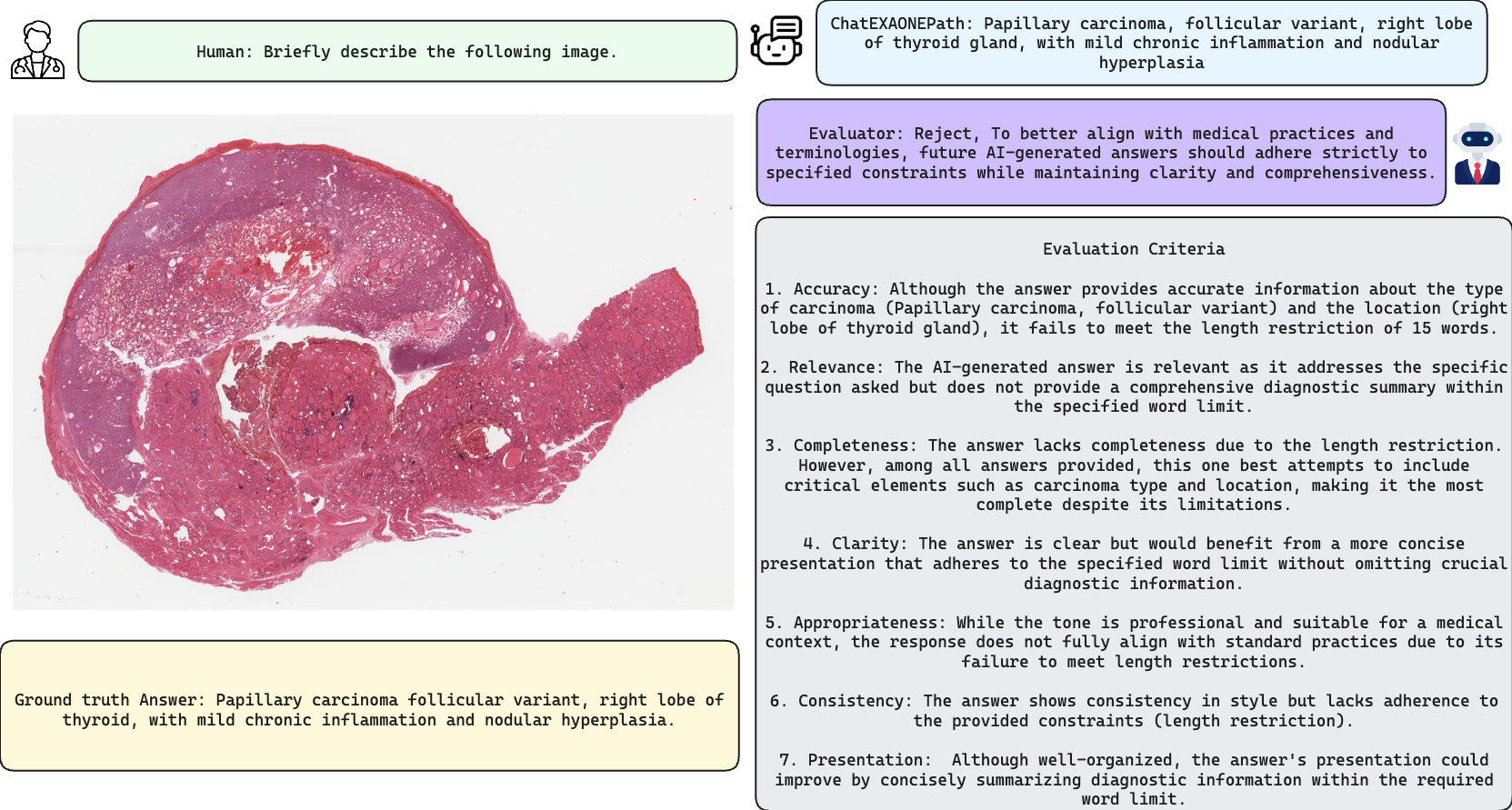}
\caption{A qualitative result and evaluation from the test dataset. ChatEXAONEPath-v3 answered from the given whole slide image and question. The evaluator AI model generated reasons (primarily length restriction of the generated answer) to \textbf{reject} the best answer, and made a final decision.}
\label{fig:result_sup_2}
\end{figure*}

\begin{figure*}[ht!]
\centering
\includegraphics[width=0.9\textwidth]{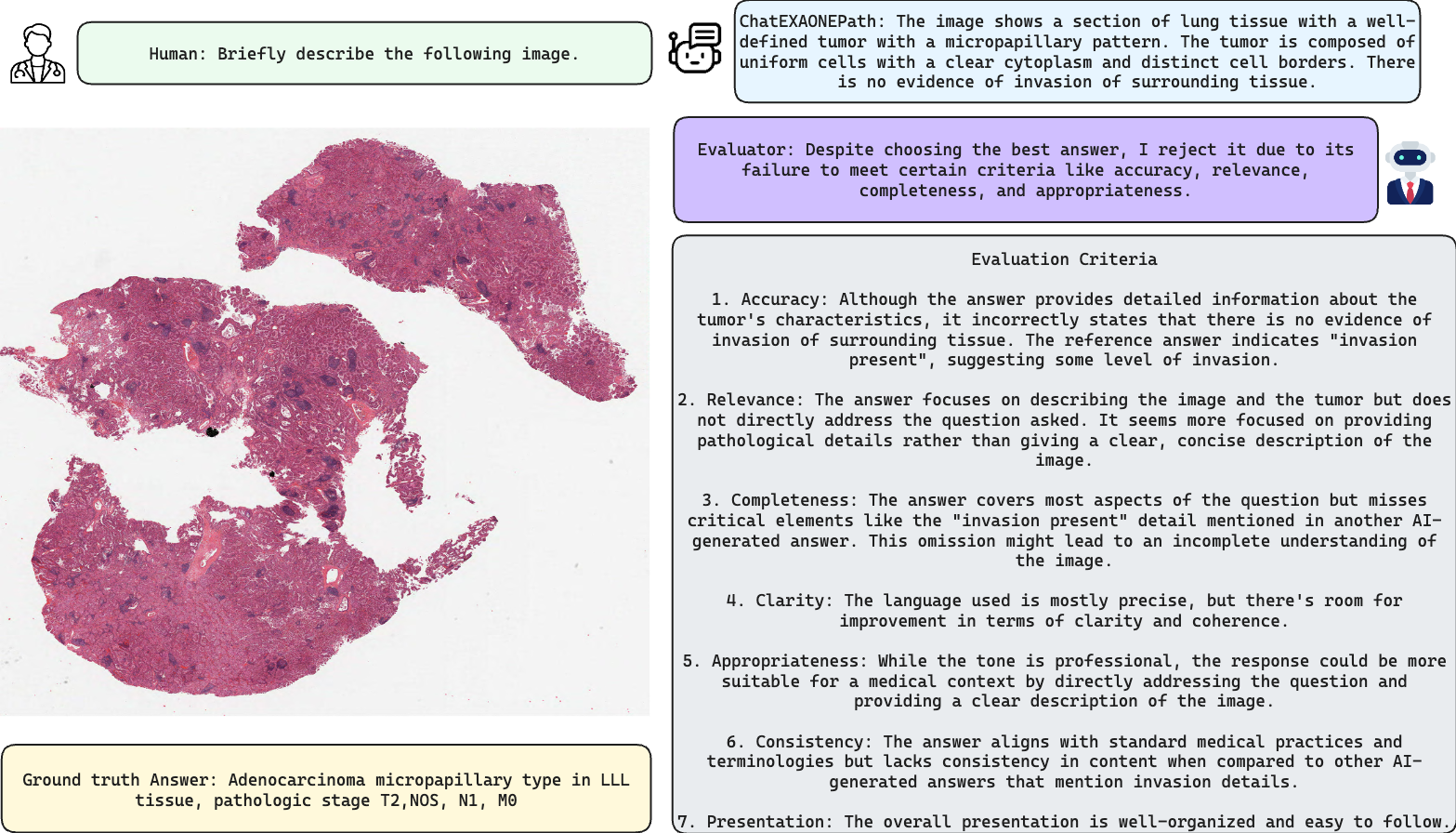}
\caption{A qualitative result and evaluation from the test dataset. ChatEXAONEPath-v1 answered from the given whole slide image and question. The evaluator AI model generated reasons (primarily length restriction of the generated answer) to \textbf{reject} the best answer, and made a final decision.}
\label{fig:result_sup_3}
\end{figure*}

In technical appendix, we introduce detailed prompts used in \textbf{R}etrieval-\textbf{A}ugmented \textbf{I}nstruction \textbf{D}ataset g\textbf{E}ne\textbf{R}ation (\textbf{RAIDER}), instruction-tuning, and AI-based evaluations in Table~\ref{table_supp_1:prompts}. In Table~\ref{table_supp_2:questions}, we present questions in the instruction-tuning datasets. In Figure~\ref{fig:result_sup_1}, Figure~\ref{fig:result_sup_2}, and Figure~\ref{fig:result_sup_3}, we showcase qualitative results of generated results from ChatEXAONEPath-v1 and ChatEXAONEPath-v2. 
\begin{table}[ht!]
\centering
\begin{tabular}{l}
\toprule
\textbf{Questions} \\
\midrule
1. What is a major diagnosis? \\
2. What is the crucial diagnosis? \\
3. What is the key diagnosis? \\
4. What is the primary diagnosis? \\
5. What is the key histopathological feature observed? \\
6. What is the main diagnosis? \\
7. What is the major diagnosis? \\
8. What is the most likely diagnosis? \\
\bottomrule
\end{tabular}
\caption{List of questions used in the instruction datasets for ChatEXAONEPath to generate the diagnosis based on the given whole slide images.}
\label{table_supp_2:questions}
\end{table}
\begin{table}[ht!]
\centering
\begin{tabular}{l|l}
\toprule
\textbf{Type} & \textbf{Prompts} \\
\midrule
\multirow{5}{1.5cm}{RAIDER} & \parbox[t]{0.85\textwidth}{You are a pathology lab assistant. You are given an unstructured pathology report describing a tissue sample of whole slide image. Follow these instructions carefully: 1. Extract a detailed summary of the diagnosis and the examined tissue from the report in a sentence under 10 words. 2. Do not mention any cm or mm measurements. 3. Do not mention any arabic or roman numerals. 4. Please give a complete and concise answer for the question. Answer the question based only on the provided context:} \\
\midrule
\multirow{7}{1.5cm}{Instruction \\ Tuning} & \parbox[t]{0.85\textwidth}{You are a specialized AI assistant in a pathology lab. Your primary role is to analyze tissue samples from images and provide concise diagnostic summaries. Please follow these instructions carefully: 1. Identify Tissue Type: If possible, specify the type of tissue visible in the image. 2. Diagnostic Summary: Provide a single-sentence diagnosis based on the image. 3. Length Restriction: Keep the summary under 20 words. 4. Exclusions: Do not include numerical measurements, units (cm, mm), or extraneous details. 5. Clarity and Brevity: Use clear, precise language and minimize unnecessary words.\\
Your response should be focused, relevant, and contain only the essential diagnostic information.} \\
\midrule
\multirow{40}{1.5cm}{Evaluation} & \parbox[t]{0.85\textwidth}{As an experienced pathologist with extensive expertise in the field, your role is to evaluate the quality of AI-generated answers in the context of medical inquiries. We aim to ensure that the AI model provides responses that are not only accurate but also contextually appropriate and professionally acceptable. Your evaluation should be based on the following detailed criteria: 
Please provide only the words “accept” or “reject” as your response.\\
            1. Accuracy: \\
            - Verify whether the AI-generated answer accurately reflects the medical facts and knowledge provided in the reference answer.\\
            - Check for any factual errors, misleading information, or incorrect interpretations that deviate from established medical knowledge.\\
            2. Relevance: \\
            - Assess whether the AI-generated answer directly addresses the specific question asked. \\
            - Ensure that the answer remains focused on the core topic without deviating into unrelated areas or including extraneous information. \\
            3. Completeness:\\
            - Evaluate whether the AI-generated answer thoroughly covers all aspects of the question as addressed in the reference answer.\\
            - Determine if any critical elements or important details are missing from the response, and whether the answer provides a comprehensive view.\\
            4. Clarity:\\
            - Judge the clarity and coherence of the AI-generated answer. \\
            - Ensure that the language used is precise, easy to understand, and free of jargon or ambiguous terms that could confuse the reader.\\
            5. Appropriateness:\\
            - Assess the professionalism and tone of the AI-generated answer. \\
            - Confirm that the response is suitable for a professional medical context, with a respectful and appropriate style for the audience.\\
            6. Consistency:\\
            - Compare the AI-generated answer with the reference answer for consistency in content and style.\\
            - Evaluate whether the AI's response aligns with standard medical practices and terminologies.\\
            7. Presentation:\\
            - Consider the overall presentation of the AI-generated answer, including formatting, grammar, and punctuation.\\
            - Ensure that the answer is well-organized and visually easy to follow, contributing to an overall polished appearance.\\
            Based on your thorough evaluation using the criteria above, please determine the acceptability of the AI-generated answer. Respond with “accept” if the AI-generated answer meets the standards of accuracy, relevance, completeness, clarity, appropriateness, consistency, and presentation. If the answer fails to meet any of these criteria or shows significant deficiencies, respond with “reject.” \\
            Question: \texttt{question}\\
            Reference Answer: \texttt{answer}\\
            AI-generated Answer: \texttt{llm answer}} \\
\bottomrule
\end{tabular}
\caption{Overview of system prompts used in Retrieval-Augmented Instruction Dataset gEneRation (RAIDER), the instruction-tuning phase of training ChatEXAONEPath, and AI-based evaluations.}
\label{table_supp_1:prompts}
\end{table}

\end{document}